\documentclass[conference]{IEEEtran}
\usepackage{url}
\usepackage{cite}
\usepackage[normalem]{ulem}
\usepackage{graphicx}
\usepackage{amsmath} 
\usepackage{booktabs}
\usepackage[ruled,vlined]{algorithm2e}
\usepackage{float}
\usepackage[dvipsnames]{xcolor}
\usepackage[]{algorithm2e}
\usepackage{algpseudocode}

\usepackage{tcolorbox}
\usepackage{listings}
\usepackage{caption}
\usepackage{subcaption}
\ifCLASSINFOpdf

\else

\fi

\SetKw{KwInit}{Initialization}

\begin{document}

\title{ Liveness Detection in Computer Vision: Transformer-based Self-Supervised Learning for Face Anti-Spoofing}

\author{\IEEEauthorblockN{Arman Keresh,
Pakizar Shamoi\IEEEauthorrefmark{1}}
\IEEEauthorblockA{School of Information Technology and Engineering \\
Kazakh-British Technical University\\
Almaty, Kazakhstan\\
Email:
\IEEEauthorrefmark{1}p.shamoi@kbtu.kz,
}
}

\maketitle

\IEEEpeerreviewmaketitle
\begin{abstract}





 Face recognition systems are increasingly used in
biometric security for convenience and effectiveness. However,
they remain vulnerable to spoofing attacks, where attackers use
photos, videos, or masks to impersonate legitimate users. This research addresses these vulnerabilities by exploring the Vision Transformer (ViT) architecture, fine-tuned with the DINO framework. The DINO framework facilitates self-supervised learning, enabling the model to learn distinguishing features from unlabeled data. We compared the performance of the proposed fine-tuned ViT model using the DINO framework against a traditional CNN model, EfficientNet b2, on the face anti-spoofing task. Numerous tests on standard
datasets show that the ViT model performs better than the
CNN model in terms of accuracy and resistance to different spoofing
methods.  Additionally, we collected our own dataset from a biometric application to validate our findings further. This study highlights the superior performance of transformer-based architecture in identifying complex spoofing cues, leading to significant advancements in biometric security.
\end{abstract}

\section{introduction}

Face recognition systems (FRS) are vital to modern security, offering efficient biometric authentication for applications like smartphone unlocking and access control \cite{frsbio}, \cite{Petrescu2019}, \cite{Nagesh2023}, \cite{fai}. These systems are particularly effective in sensitive areas, where they can restrict unauthorized access and enhance reliability \cite{Chowdhry2013}. Smartphone-based FRS are also being explored, focusing on feature extraction algorithms and security challenges \cite{AbdElaziz2021}. However, they are vulnerable to spoofing attacks, where impostors use photos, videos, or masks to mimic legitimate users and deceive the system \cite{ErdogmusNesli2014Spoofing}, \cite{Hamdan2018detection}.  Even simple identity spoofing methods, such as using mobile camera shots or social media photos, can compromise the security of these systems \cite{Omar2015Evaluating}. This vulnerability necessitates the development of robust anti-spoofing techniques to accurately distinguish between genuine and spoofed faces \cite{Omar2016Designing}.





Several recent studies have demonstrated the potential of vision transformers in face anti-spoofing \cite{Latifah2023Anomaly}, \cite{Chen2023Domain}, \cite{Liu2023}, \cite{Liao2023Domain}, \cite{Lee2023Robust}. Current research addresses this problem using the Vision Transformer (ViT) architecture, fine-tuned with the DINO (Emerging Properties in Self-Supervised Vision Transformers) framework. The DINO framework facilitates self-supervised learning, enabling the model to learn distinguishing features from unlabeled data. We hypothesize that a transformer-based model, trained on a large, diverse dataset, can effectively capture the nuanced features indicative of spoofing, thus outperforming traditional CNN models.

In this study, we utilized multiple benchmark datasets to evaluate the performance of our proposed Vision Transformer (ViT) model, fine-tuned using the DINO framework. Besides these established datasets, we also gathered a unique dataset from a biometric application.

The contributions of this study are as follows:
\begin{itemize}
    \item Introducing the Vision Transformer (ViT) architecture fine-tuned with the DINO (Emerging Properties in Self-Supervised Vision Transformers) framework for face anti-spoofing.
    \item Comparative Analysis with Traditional CNN Model, EfficientNet b2, and fine-tuned ViT model on the face anti-spoofing task. 
\end{itemize}

The paper is structured as follows: Section I is this introduction. Section II presents an overview of the works related to face anti-spoofing. Next, Section III describes the methods employed in this study, including data collection, vision transformers, and the DINO framework. Experimental results are presented in Section IV, followed by a Discussion in Section V and Future Works in Section VI. Finally, the concluding remarks are drawn in Section VII.

\section{Related Work}



In this section, we review the existing methods for face anti-spoofing, including traditional and deep learning-based approaches. The vulnerability of face recognition systems to spoofing attacks has been extensively studied.

Initial methods for face anti-spoofing mainly used handcrafted features and traditional machine learning techniques. For instance, some researchers \cite{7748511} utilized SURF - speeded-up robust features as a patented local feature detector and descriptor, and Fisher vector encoding is an image feature encoding and quantization technique to enhance face spoof detection, but these methods struggled with generalizing to new and unseen spoofing attacks. Similarly, researchers focused on smartphone-based face unlock systems, emphasizing the limitations of these traditional methods in dynamic and varied attack scenarios \cite{7487030}.

A range of other methods have been proposed for face anti-spoofing, including Haralick texture features \cite{Agarwal2016Face}, image quality assessment \cite{Fourati2017Face}, patch and depth-based CNNs \cite{Atoum2017Face}, and multifeature videolet aggregation \cite{Siddiqui2016}. These methods have shown promising results in distinguishing between genuine and spoofed face appearances. Other approaches include general image quality assessment \cite{Galbally2014Face}, color texture analysis \cite{Boulkenafet2015Face}, and pulse detection from face videos \cite{XiaobaiLi2016}, all of which have demonstrated effectiveness in detecting various types of spoofing attacks. Combining FRS with other security systems, such as RFID, has also been suggested to strengthen security \cite{Aff2013RFIDAF}.

Since the emergence of deep learning, Convolutional Neural Networks (CNNs) have become popular in face anti-spoofing research. Several studies have demonstrated the effectiveness of CNNs in learning features directly from data \cite{9070853, 9112543}, leading to improved liveness detection performance. However, these models require large, diverse datasets and often struggle with generalization to novel spoofing techniques due to their reliance on local feature extraction.

Several recent studies have explored the use of transformer architectures in face anti-spoofing, with promising results. Studies \cite{Wang2022Face} and \cite{Latifah2023Anomaly} both achieved competitive performance using ViT transformers, with the latter introducing a relation-aware mechanism.  The performance was further improved by deepening the transformer network loop depth and introducing adaptive transformers for robust cross-domain face anti-spoofing, respectively \cite{Tianqi2021Transformer}, \cite{Hsin2022Adaptive}. Other similar studies focused on generalizability, with the former proposing a domain-invariant vision transformer and the latter demonstrating the effectiveness of vision transformers for zero-shot face anti-spoofing \cite{George2021}, \cite{Chen2023Domain}.  The other work presents UDG-FAS, the first Unsupervised Domain Generalization framework for Face Anti-Spoofing \cite{unsuperv}. This framework uses large volumes of unlabeled data to learn generalizable features, thereby improving performance in low-data scenarios for face anti-spoofing. Another study introduces FM-ViT, a transformer-based framework that outperforms existing single-modal frameworks \cite{Liu2023}. Adaptive vision transformers for robust few-shot cross-domain face anti-spoofing was proposed in the other recent study \cite{huang}.  The generalizability of vision transformers was further improved with the Domain-invariant Vision Transformer (DiVT) \cite{Liao2023Domain}. Next, the study \cite{Lee2023Robust} developed a convolutional vision transformer-based framework for robust performance against unseen domain data.

As we see, recent advancements in Vision Transformers (ViTs) offer a promising alternative. Unlike CNNs, ViTs capture global dependencies via self-attention mechanisms, potentially enhancing their ability to identify subtle, global spoofing cues. Studies \cite{9484333} have explored the application of ViTs for unseen face anti-spoofing, showcasing their potential in handling unseen attacks. Further \cite{9817442} research highlighted the effectiveness of transformers in incorporating relation-aware mechanisms for improved spoof detection.



Specific challenges frequently arise in face anti-spoofing research, including difficulties in generalizing across different domains and datasets, the constraints imposed by limited data, and technical obstacles related to methodologies such as anomaly detection and the use of black-box discriminators. Cross-domain face anti-spoofing, such as the domain gap and limited data, can lead to poor generalization of models to new domains. Furthermore, the generalization capabilities of classifiers, particularly when applied to diverse databases, are often questioned, as they may not consistently perform well across different datasets.

\section{Methods}

\subsection{Data}
In this research, we employed several benchmark datasets to assess how well our proposed Vision Transformer (ViT) model fine-tuned with the DINO framework. These datasets are chosen for their diversity and comprehensive coverage of various spoofing techniques, ensuring a robust assessment of the model's capabilities.


The CelebA-Spoof \cite{zhang2020celebaspoof} dataset is a big dataset created especially for face anti-spoofing tasks. It contains over 625,000 images of 10,000 subjects, incorporating various spoofing attacks, including printed photos, replayed videos, and 3D masks. The dataset's extensive range of spoofing techniques and high subject diversity make it an excellent resource for training and evaluating anti-spoofing models, ensuring they can generalize well to different types of attacks.

The CASIA-SURF \cite{zhang2020casiasurf} dataset includes 21,000 images captured in multiple modalities: RGB, Depth, and Infrared. This multi-modal approach provides rich information that can be leveraged by deep learning models to improve spoof detection accuracy. The dataset is particularly useful for evaluating the effectiveness of models in scenarios where different types of image data are available, enhancing the robustness of anti-spoofing systems.



In addition to these well-known public datasets, we used a proprietary dataset, which we collected from a biometrics application; it is owned and controlled by a company. This dataset was created during sessions flagged as suspicious and non-suspicious. During biometric authentication, subjects were often asked to turn their heads or move closer, resulting in a dataset of 100,000 images. Each subject underwent multiple biometric sessions, providing diverse images under various conditions. These images are unlabeled. Due to privacy concerns and the sensitive nature of the biometric data, this dataset cannot be publicly disclosed. Using a self-supervised learning approach, we aim to train a Vision Transformer (ViT) on this unlabeled data.

\begin{figure}
     \centering
     \begin{subfigure}[b]{0.23\textwidth}
         \centering
         \includegraphics[width=\textwidth]{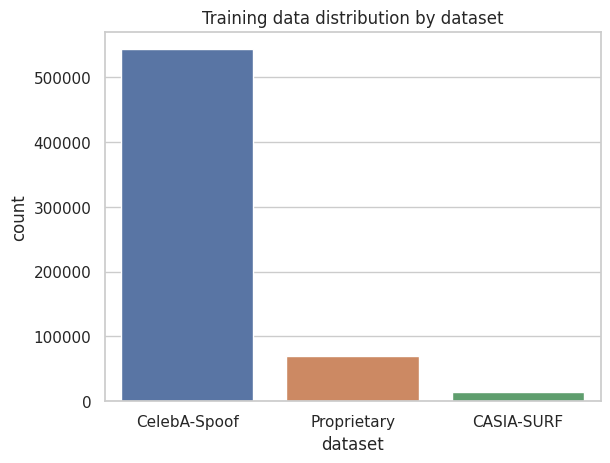}
         \caption{Dataset distribution}
         \label{fig:y equals x}
     \end{subfigure}
     \hfill
     \begin{subfigure}[b]{0.23\textwidth}
         \centering
         \includegraphics[width=\textwidth]{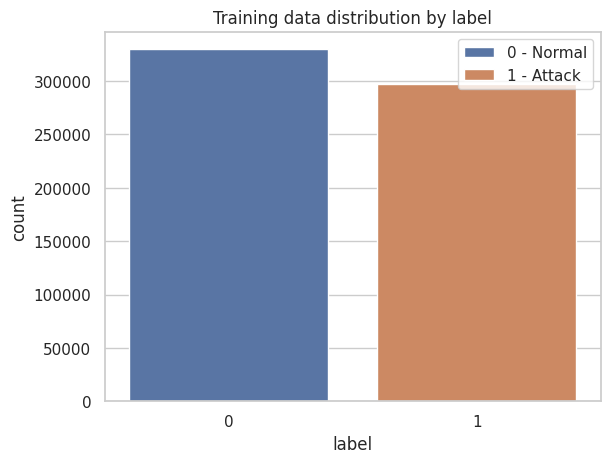}
         \caption{Label distribution}
         \label{fig:three sin x}
     \end{subfigure}
        \caption{Trainig data distribution by dataset and label}
        \label{fig:train data distributions}
\end{figure}

\begin{figure}
     \centering
     \begin{subfigure}[b]{0.23\textwidth}
         \centering
         \includegraphics[width=\textwidth]{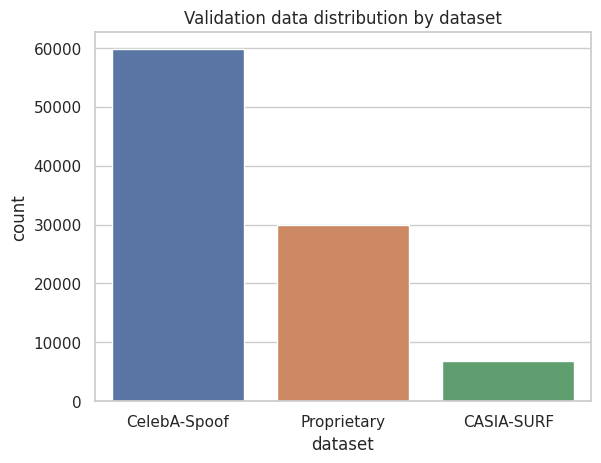}
         \caption{Dataset distribution}
         \label{fig:y equals x}
     \end{subfigure}
     \hfill
     \begin{subfigure}[b]{0.23\textwidth}
         \centering
         \includegraphics[width=\textwidth]{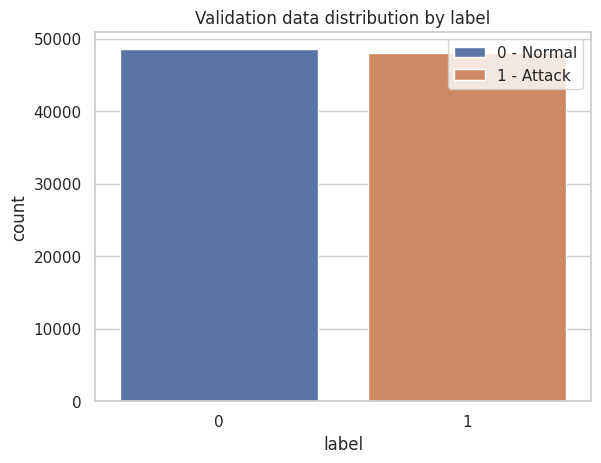}
         \caption{Label distribution}
         \label{fig:three sin x}
     \end{subfigure}
        \caption{Validation  data distribution by dataset and label}
        \label{fig:validation data distributions}
\end{figure}

The dataset used in this study consists of images from three sources: CelebA-Spoof, a proprietary dataset, and CASIA-SURF. The training data distribution, as depicted in the first set of plots (see Figure \ref{fig:train data distributions}), shows that the majority of the data comes from the CelebA-Spoof dataset with 543,424 images, followed by the proprietary dataset with 69,234 images, and CASIA-SURF with 14,879 images (Table \ref{tab:data_distribution}). For the validation data, the distribution is similar, with 59,762 images from CelebA-Spoof, 29,856 images from the proprietary dataset, and 6,892 images from CASIA-SURF. (see Figure \ref{fig:validation data distributions}) These distributions highlight the reliance on the CelebA-Spoof dataset for training and validation, supplemented by the proprietary and CASIA-SURF datasets to provide a diverse set of images for evaluating the model's performance across different sources. This diverse dataset composition is crucial for ensuring the robustness and generalizability of the face anti-spoofing models developed in this study. The label distribution (Table \ref{tab:label_distribution}) also indicates a balanced representation of normal and attack labels in both training and validation sets, which is essential for accurate model training and evaluation.

\begin{table}[H]
\centering
\begin{tabular}{|c|c|c|}
\hline
\textbf{Dataset} & \textbf{Split} & \textbf{Total}  \\
\hline
CelebA-Spoof & Train & 543424 \\
CelebA-Spoof & Validation & 59762 \\
Proprietary & Train & 69234  \\
Proprietary & Validation & 29856\\
CASIA-SURF & Train & 14879 \\
CASIA-SURF & Validation & 6892 \\
\hline
\end{tabular}
\caption{Distribution of data in train and validation sets}
\label{tab:data_distribution}
\end{table}

\begin{table}[h]
\centering
\begin{tabular}{|c|c|c|}
\hline
\textbf{Split} & \textbf{Label 0 (Normal)} & \textbf{Label 1 (Attack)}  \\
\hline
Train & 329850 & 297687 \\
Validation & 48520 & 47990 \\
\hline
\end{tabular}
\caption{Distribution of labels in train and validation sets}
\label{tab:label_distribution}
\end{table}

\begin{figure}[h]
    \centering
    \includegraphics[width=0.5\textwidth]{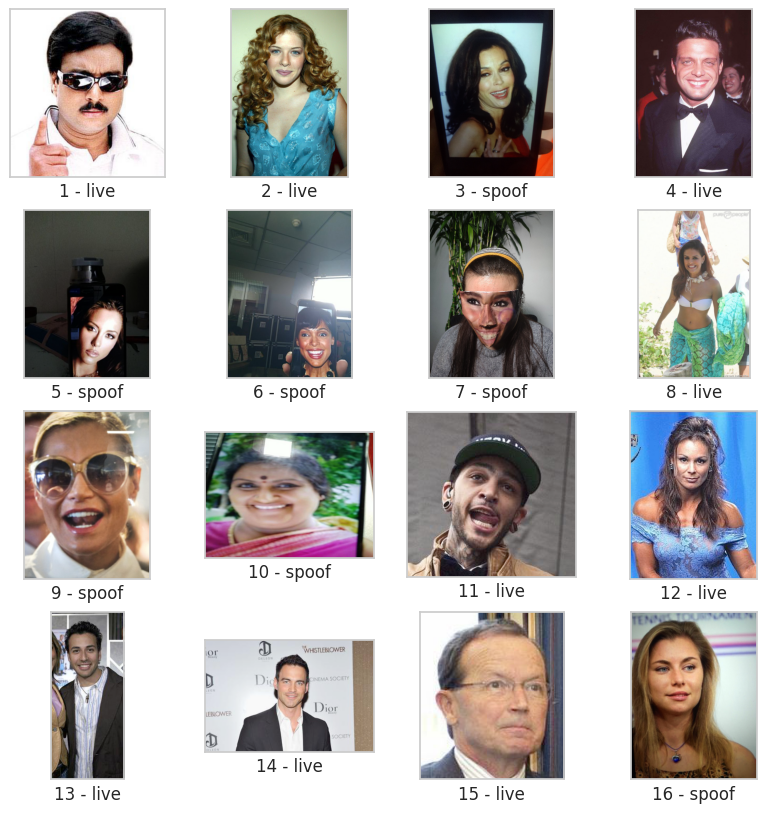}
    \caption{Sample images from the dataset, illustrating various genuine (“live”) and fake (“fake”) examples. The dataset includes a variety of facial images, including spoofing techniques such as printed photographs, screen images.}
    \label{fig:my_label}
\end{figure}

Figure \ref{fig:my_label} shows a sample of images from the dataset used in this study. The dataset includes a diverse range of face images, both genuine and spoofed, to train and evaluate the face anti-spoofing models. The images in the sample illustrate various spoofing techniques, such as printed photos (images 5-7), screen displays (images 3, 9-10), and genuine face images. Each image is labeled as either "live" or "spoof," highlighting the ground truth for training and validation purposes.



\subsection{Vision Transformer (ViT)}
\begin{figure*}[bt]
    \centering
    \includegraphics[width=0.8\linewidth]{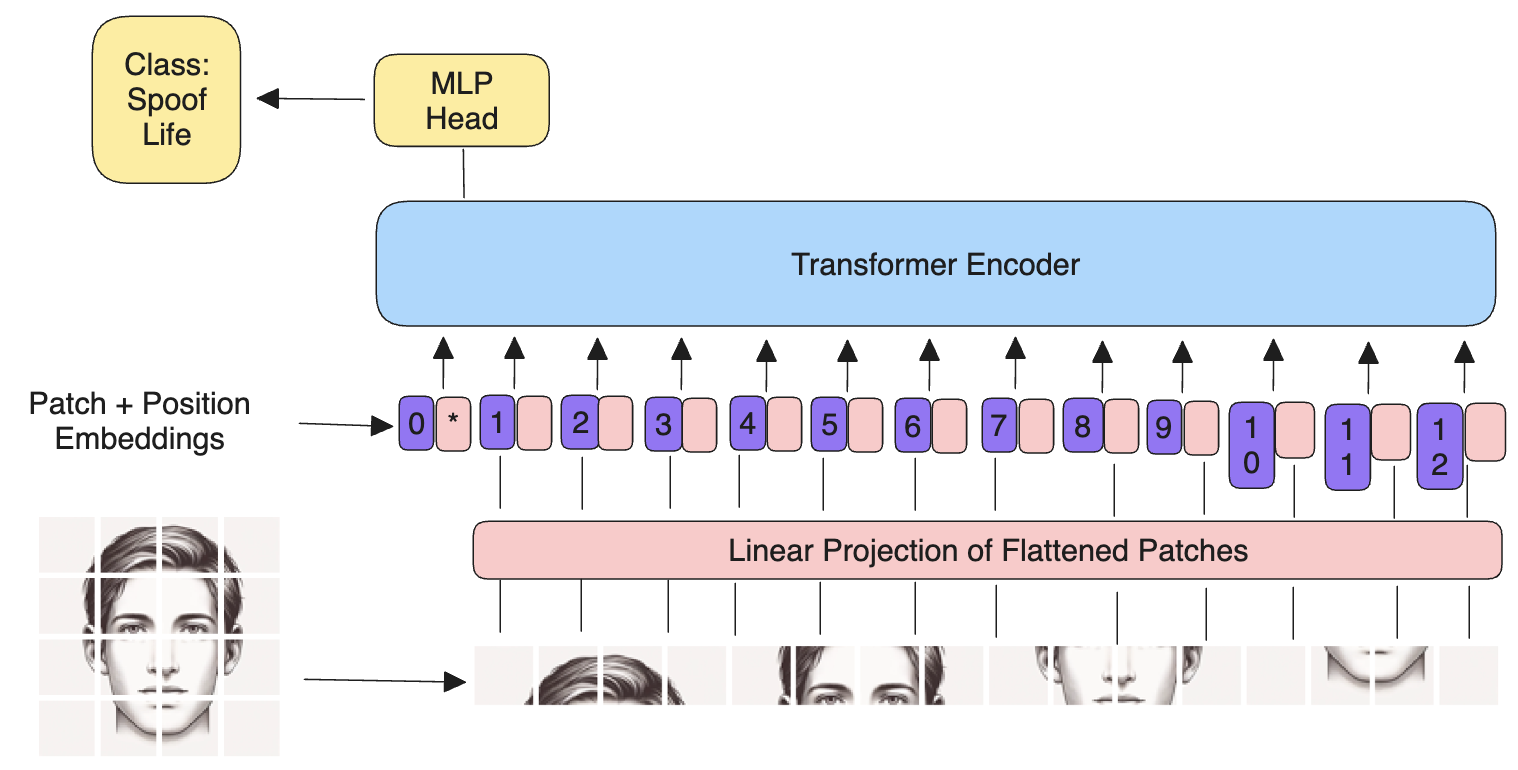}
    \caption{The input face image is split into patches, which are then projected linearly and embedded with positional information. These embeddings go into the Transformer encoder, which processes the sequence of patches. The encoder's output is then passed through a multi-layer perceptron (MLP) head to classify the image as either "spoof" or "live." 
    }
    \label{fig:vit}
\end{figure*}

Vision Transformers significantly impacted the field of computer vision \cite{fai2}. Vision Transformer (ViT) architecture treats an image as a sequence of patches, similar to how words are treated in text processing using Transformers  \cite{Dosovitskiy2020}. Each image is split into a grid of non-overlapping patches, then linearly embedded and provided with positional embeddings. These embeddings go through a standard Transformer encoder, which uses multi-head self-attention mechanisms to understand the connections between different patches (see Fig. \ref{fig:vit}).

The self-attention mechanism in Transformers can be defined as:

\begin{equation}
\text{Attention}(Q, K, V) = \text{softmax}\left(\frac{QK^T}{\sqrt{d_k}}\right)V
\end{equation}

where \( Q \) (queries), \( K \) (keys), and \( V \) (values) are derived from the input embeddings, and \( d_k \) is the dimension of the keys.

\subsection{DINO (Distillation with No Labels)}
\begin{figure*}[t]
    \centering
    \includegraphics[width=\textwidth]{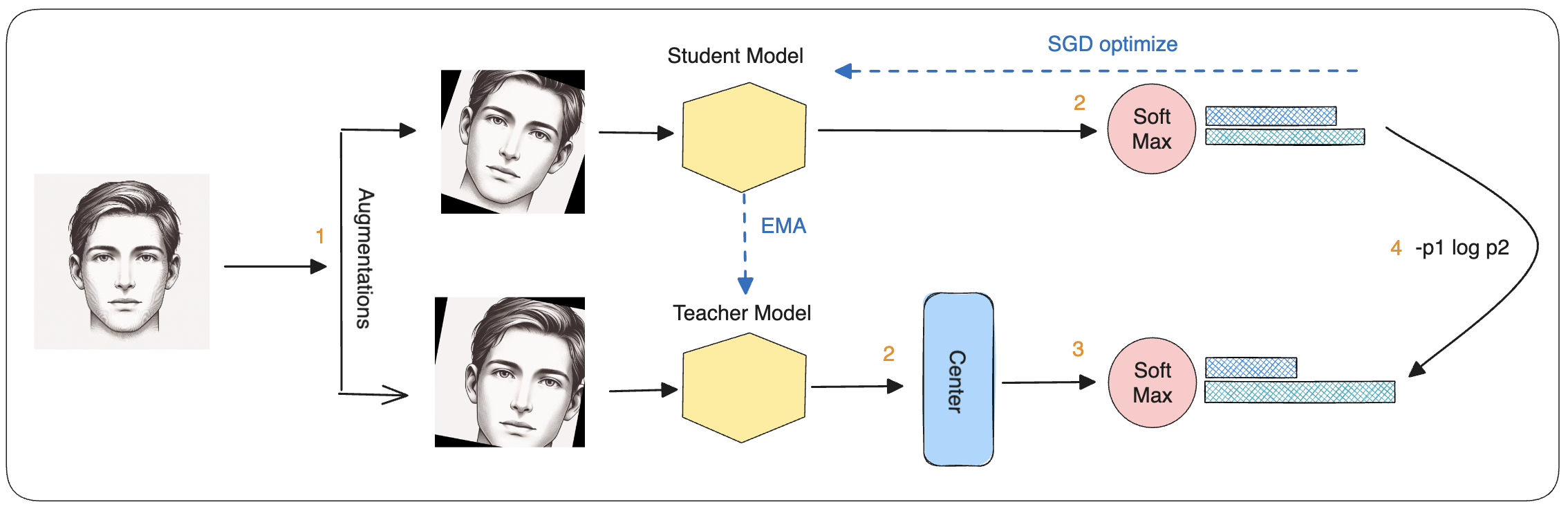} 
    \caption{This figure illustrates the DINO (Distillation with No Labels) model training process. It starts with image augmentations (1), where two augmented views of the same image are generated. The student model processes one view, while the teacher model processes the other (2). The teacher model's outputs are centered and passed through a softmax layer (3). The student's outputs are optimized using Stochastic Gradient Descent (SGD) to match the teacher's outputs via an exponential moving average (EMA) update (4), minimizing the cross-entropy loss between the student's and teacher's predictions. }
    \label{fig:dino model}
\end{figure*}

DINO is a self-supervised learning approach that trains the model to generate similar embeddings for different views of the same image \cite{Caron2021}. This is done using a student-teacher training setup, where the student network learns to imitate the output of the teacher network. Architecture is shown in Fig. \ref{fig:dino model}. 

\begin{itemize}
    \item Teacher Network: A fixed pre-trained network that provides stable target representations.
    \item Student Network: A trainable network that learns to predict the teacher's representations.
\end{itemize}

The DINO framework helps the ViT model learn discriminative features from large amounts of unlabeled data. This is particularly useful for tasks like face anti-spoofing, where labeled data may be limited. It will help the model train on our data without labels.


\subsection{EfficientNet b2}
EfficientNet b2 is a CNN model optimized for both efficiency and performance \cite{tan2020efficientnet}. 
It uses a compound scaling method that proportionally increases the network's width, depth, and resolution, resulting in improved accuracy with fewer parameters and reduced computational cost. To enhance its performance further, we employed the noisy student \cite{xie2020selftraining} training approach, which iteratively trains the model on our custom unlabeled dataset, leveraging self-training with noise to improve robustness and accuracy. For training, we utilized the CelebA-Spoof and CASIA-SURF datasets. Additionally, our custom dataset, consisting of 1,000,000 images from biometric applications, was incorporated into the training process using the noisy student approach, enhancing the model's ability to generalize across different spoofing scenarios.

\subsection{Proposed Approach}

To tackle the issue of face anti-spoofing, we fine-tuned a Vision Transformer (ViT) model using the DINO framework.  Our approach leverages ViTs' ability to capture global dependencies in the input data via self-attention mechanisms, which enhances their ability to detect subtle, global spoofing cues. We compared how well the ViT model performed against a traditional Convolutional Neural Network (CNN) model, EfficientNet b2, to see how effective transformer-based methods are in this field. Our models were trained on two NVIDIA A100 40 GB GPUs. The detailed training procedure is outlined in Algorithm \ref{algo}.

During the training process, a variety of data augmentation techniques were used to enhance the robustness and generalizability of the face anti-spoofing models. These augmentations were categorized into four main groups: 
\begin{enumerate}
    \item \textbf{Color Transformations}. To provide color variations and simulate different lighting conditions, we used augmentations such as ChannelShuffle, ChannelDropout, RandomBrightnessContrast.
    \item \textbf{Affine Transformations}. To provide geometric variations and enhance the model's ability to generalize across different orientations and perspectives, we used augmentations such as Rotate, Flip. 
    \item \textbf{Quality Degradations}. To simulate various image quality issues that might be encountered in real-world scenarios, we used augmentations such as ImageCompression, and a combination of blurring techniques such as Blur with a blur limit of 3 to 7, MotionBlur with a blur limit of 7 to 20, GaussNoise for variability in noise levels.
    \item \textbf{Cropping and Padding}. To alter the spatial composition of the images, we used CropAndPad with a percentage range of -10\% to +23\% which randomly crops and pads the images, ensuring the model can handle partial occlusions and varying framing conditions.
\end{enumerate}

\begin{algorithm}
    \caption{Training DINOv2 for Liveness and Anti-Spoofing Classification}
    \label{algo}
    \SetAlgoLined
    \KwIn{Dataset $D$ (CelebA-Spoof, CASIA-SURF, Proprietary), Image Size $260 \times 260$}
    \textbf{Initialization:}{ DINOv2 model $M$ with pre-trained weights, with Batch Size $B$ = $4$, Learning Rate $LR$ $0.001$}\;
    \KwOut{ViT model $DINOv2$}
    \For{epoch = 1 \KwTo 300}{
        \For{each batch $B$ in $D$}{
            Resize images in $B$ to $260 \times 260$\;
            Apply augmentations to images\;
            Forward pass through $M$ with half-precision (fp16)\;
            Compute FocalLoss on predictions\;
            Update $M$ using Adam optimizer\;
            Adjust $LR$ with OneCycleLR scheduler\;
            }
    }
    \Return Trained model $M$
\end{algorithm}

\textbf{Training Algorithm:}
\begin{enumerate}
    \item \textbf{Data Preparation}.Split images into patches and create patch embeddings with positional encodings.
    \item \textbf{Self-Supervised Pre-training}. Use the DINO framework to pre-train the ViT model on a large dataset of unlabeled facial images.
    \item \textbf{Fine-tuning}. Replace the decoder with a binary classification layer and fine-tune the model on labeled face anti-spoofing datasets.
    \item \textbf{Evaluation}. Compare the performance of the ViT model with EfficientNet b2 using standard metrics.
\end{enumerate}

\begin{figure*}[t]
    \centering
    \includegraphics[width=\textwidth]{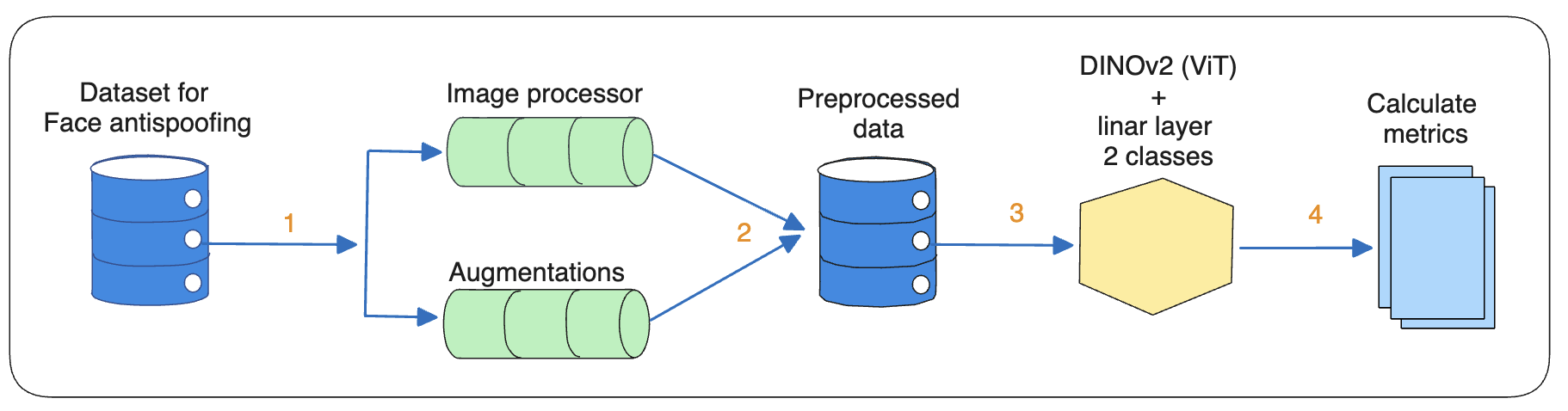} 
    \caption{The training pipeline involves collecting a comprehensive dataset for face anti-spoofing, processing and augmenting the images to enhance quality and variability, feeding the preprocessed data into the DINOv2 (Vision Transformer) model with a binary classification layer, and evaluating the model's performance. }
    \label{fig:training_pipeline}
\end{figure*}

See Fig. \ref{fig:training_pipeline} for the detailed training algorithm steps.


\section{Experimental Results}

To evaluate the performance of the models, we used standard metrics in face anti-spoofing, including APCER, BPCER, and ACER. \cite{ISO30107-3}
We express APCER and BPCER in terms of true positives (TP), false positives (FP), true negatives (TN), and false negatives (FN)

\textbf{APCER} (Attack Presentation Classification Error Rate): This is the rate of attack presentations (spoof attempts) incorrectly classified as bona fide (genuine) presentations. 

\begin{equation}
\text{APCER} = \frac{\text{FP}}{\text{FP} + \text{TN}}
\end{equation}

\textbf{BPCER} (Bona Fide Presentation Classification Error Rate): This is the rate of bona fide presentations incorrectly classified as attack presentations.

\begin{equation}
\text{BPCER} = \frac{\text{FN}}{\text{FN} + \text{TP}}
\end{equation}

\textbf{ACER} (Average Classification Error Rate): This is the mean of APCER and BPCER, providing a single metric to evaluate the model's overall performance.

\begin{equation}
\text{ACER} = \frac{\text{APCER} + \text{BPCER}}{2}
\end{equation}

\begin{figure}[bt]
    \centering
    \includegraphics[width=\linewidth]{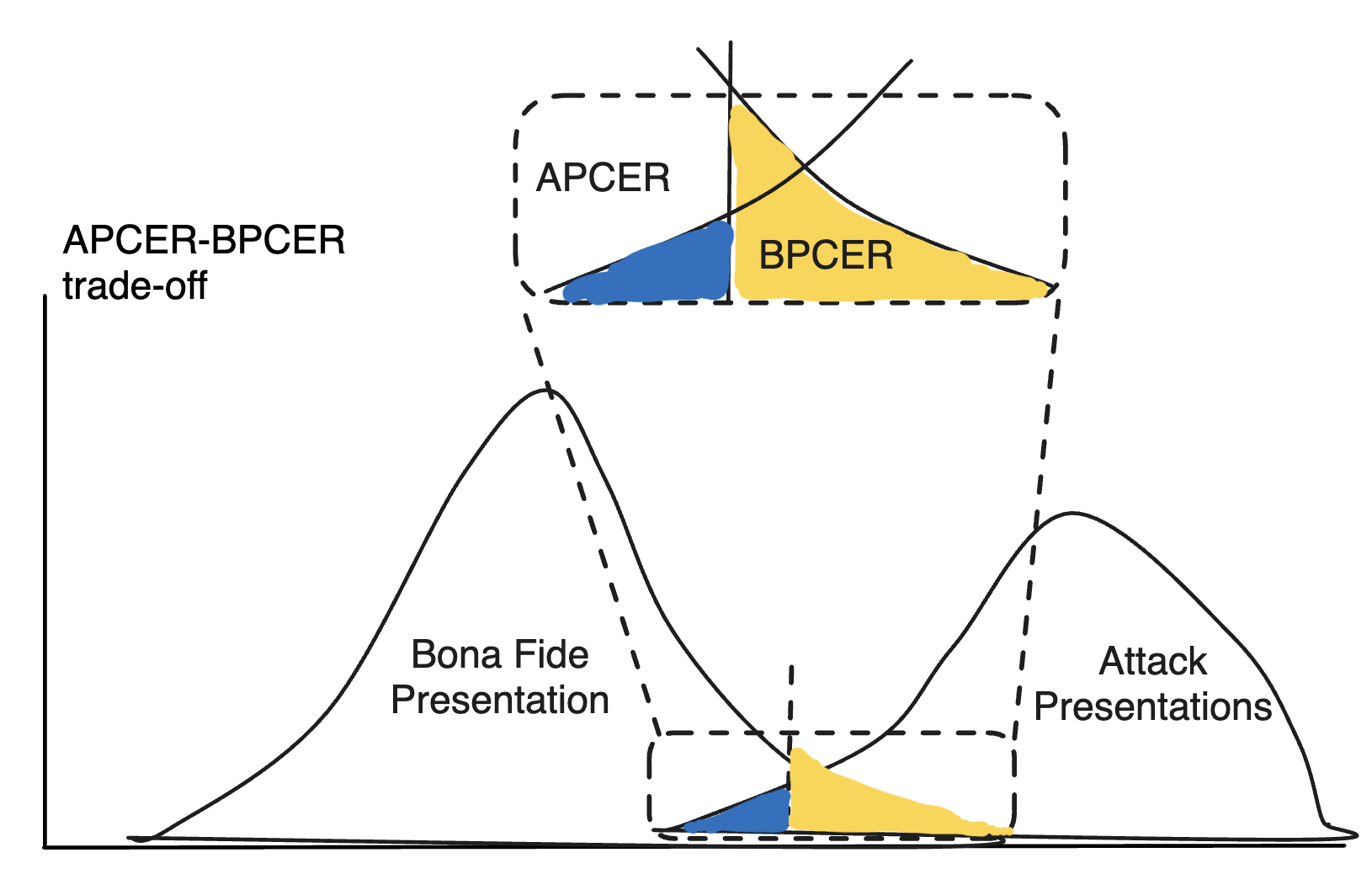}
    \caption{Balancing the Attack Presentation Classification Error Rate (APCER) and the Bona Fide Presentation Classification Error Rate (BPCER) is important. The overlapping areas show misclassifications: APCER (blue) represents attack presentations wrongly classified as real, and BPCER (yellow) shows real presentations wrongly classified as attacks. Finding the right balance between these two metrics is key to improving the performance of face anti-spoofing systems. }
    \label{fig:apcer_bpcer}
\end{figure}

In face anti-spoofing systems, APCER and BPCER present a trade-off Fig. \ref{fig:apcer_bpcer}. Minimizing APCER (reducing false acceptance of spoofs) can increase BPCER (false rejection of genuine attempts) and vice versa. Balancing these rates is crucial for effective performance.

The performance metrics for both models are summarized in Table \ref{table:comparison}. The results demonstrate that the ViT (DINO) model significantly outperforms the EfficientNet b2 model across all evaluation metrics. Table \ref{table:comparison_vit_effnet_datasets} provide a detailed comparison of the models performance on different datasets. 

\begin{table*}[bt]
\centering
\caption{Comparison of EfficientNet and ViT (DINO) Models}
\begin{tabular}{lcc}
\toprule
\textbf{Metric} & \textbf{EfficientNet b2} & \textbf{ViT (DINO)} \\
\midrule
APCER      & 16.7 & 1.6 \\
BPCER      & 0.7  & 0.1 \\
ACER       & 8.7  & 0.8 \\
Accuracy (\%)  & 99.1 & 99.8 \\
\bottomrule
\end{tabular}
\label{table:comparison}
\end{table*}

\begin{table*}[bt]
\centering
\caption{Comparison of ViT (DINO) and EfficientNet b2 models on different datasets} 
\begin{tabular}{lcccccc}
\toprule
\textbf{Metric} & \multicolumn{3}{c}{\textbf{ViT (DINO) Model}} & \multicolumn{3}{c}{\textbf{EfficientNet b2 Model}} \\
\cmidrule(r){2-4} \cmidrule(r){5-7}
 & \textbf{CelebA-Spoof} & \textbf{CASIA-SURF} & \textbf{Proprietary} & \textbf{CelebA-Spoof} & \textbf{CASIA-SURF} & \textbf{Proprietary}\\
\midrule
APCER      & 2.97 & 1.7 & 0.12 & 27.37 & 12.53 & 8.22 \\
BPCER      & 0.11  & 0.0032 & 0.186 & 0.89  & 0.7 & 0.51 \\
ACER       & 1.54  & 0.85 & 0.15 & 14.13  & 6.63 & 4.365 \\
Accuracy (\%)  & 99.5 & 99.8 & 99.9 & 98.52 & 98.97 & 99.78 \\
\bottomrule
\end{tabular}
\label{table:comparison_vit_effnet_datasets}
\end{table*}

The ViT (DINO) model significantly outperforms the EfficientNet b2 model across all metrics. It achieved a lower APCER (1.6\% vs. 16.7\%) and BPCER (0.1\% vs. 0.7\%), indicating better performance in identifying both attack and bona fide presentations. The ACER for ViT (DINO) was also lower (0.8\% vs. 8.7\%), demonstrating a better balance in handling both types of presentations. Additionally, the ViT (DINO) model achieved higher overall accuracy (99.8\% vs. 99.1\%), highlighting its superior ability to distinguish genuine faces from spoofed ones.

\begin{figure}[tb]
    \centering
    \includegraphics[width=\linewidth]{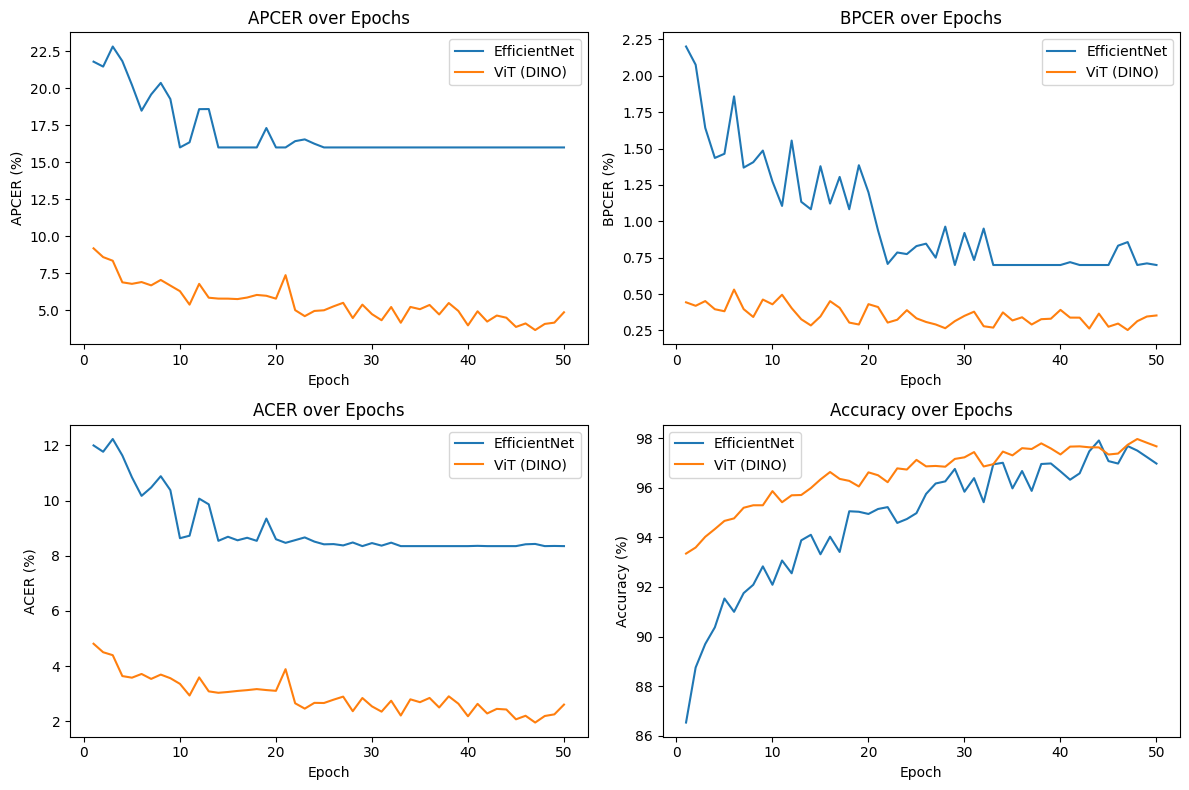}
    \caption{Trends of APCER, BPCER, ACER, and accuracy over 50 training epochs for EfficientNet b2 and ViT (DINO) models, demonstrating the superior performance of the ViT (DINO) model in face anti-spoofing tasks. } 
    \label{fig:metrics}
\end{figure}

Fig. \ref{fig:metrics} illustrates the trends for APCER, BPCER, ACER, and accuracy over 50 training epochs for both models. The plot demonstrates a significant decrease in APCER for both models, with the ViT (DINO) model consistently maintaining a lower APCER throughout the training process. The BPCER plot highlights the reduction in BPCER, where the ViT (DINO) model shows superior performance by achieving a lower BPCER than EfficientNet b2. The ACER plot indicates the overall classification error rates, significantly improving the ViT (DINO) model's ability to balance APCER and BPCER. The accuracy plot illustrates the higher overall accuracy of the ViT (DINO) model compared to EfficientNet b2, indicating better general performance in distinguishing genuine and spoofed faces.


\section{Discussion}

The ViT (DINO) model's superior performance can be attributed to its ability to capture global dependencies and subtle spoofing cues through self-attention mechanisms. The DINO framework's self-supervised pre-training further enhances the model's capability to learn discriminative features from large amounts of unlabeled data, improving its robustness and generalization to diverse spoofing techniques.

In contrast, the EfficientNet b2 model, despite its optimized architecture for efficiency and performance, relies on local feature extraction through convolutional layers, which may limit its ability to generalize to novel and sophisticated spoofing attacks. Additionally, the traditional supervised learning approach used for training EfficientNet b2 may not exploit the full potential of the available data as effectively as the self-supervised approach used for the ViT model.

The findings of this study suggest that adopting transformer-based architectures, such as ViT, fine-tuned with self-supervised learning frameworks like DINO, can significantly enhance the performance of face anti-spoofing systems. This progress has practical benefits for enhancing the security and reliability of biometric authentication systems, which are used increasingly in different areas, such as unlocking personal devices and controlling access in secure places.


Let's review how the current study's results compare to those of previous studies. A range of studies have explored the use of vision transformers in face anti-spoofing, with promising results. 

Many studies demonstrate the effectiveness of these models in detecting anomalies and achieving robust performance across different domains \cite{Liao2023Domain}, \cite{dis1}, \cite{Latifah2023Anomaly}, \cite{Huang2022Adaptive}. Studies \cite{9817442} and \cite{Yang2024} further enhance the capabilities of vision transformers by incorporating relation-aware mechanisms and adaptive-avg-pooling-based attention. Next, \cite{George2021On} and \cite{Orfao2023} extend the application of vision transformers to zero-shot anti-spoofing and data augmentation, respectively, also achieving state-of-the-art performance. Lastly, \cite{Tianqi2021Transformer} and \cite{Watanabe2022} both report significant improvements in accuracy and reduced equal error rates using transformer-based models.  These studies collectively highlight the potential of vision transformers in enhancing the security of face recognition systems. Our findings back up these prior research research works. 

Although similar research had previously been carried out, fine-tuning the ViT architecture with Dino has received little attention in the literature.

\section{Limatations and Future Works}


Study has certain limitations to be addressed in future work. Firstly, the reliance on a specific set of datasets may limit the generalizability of the results to other types of spoofing attacks or different demographic groups. Secondly, while the DINO framework provides significant improvements, it also introduces additional computational complexity that may be challenging to implement in real-time applications. Finally, the current study does not consider the potential impact of environmental variations, such as lighting conditions and camera quality, on the model's performance. Addressing these limitations in future research will be crucial for developing more universally applicable and efficient face anti-spoofing systems.

Future research should consider using extra data types, like depth and infrared, to make face anti-spoofing models even more robust. Investigating the application of other self-supervised learning techniques and transformer architectures could also provide further enhancements. In addition, in future research, we aim to explore the integration of fuzzy logic with ViT, a recent trend \cite{fan2024vitar}, \cite{fuzz-ie}. Fuzzy logic is a powerful tool for handling imprecision and uncertainty \cite{Kozlov2024}, which could enhance the robustness and adaptability of face anti-spoofing models, particularly in scenarios with ambiguous or uncertain data. Finally, real-world testing and deployment of these models in diverse environments would be valuable in assessing their practical effectiveness and identifying areas for improvement.

\section{Conclusion}


 In this paper, we investigated the Vision Transformer architecture, which has been fine-tuned using the DINO framework to handle the face anti-spoofing problem. Several benchmark datasets were used to assess the effectiveness of the model.

The results of this study demonstrate the significant advantages of using Vision Transformer (ViT) models fine-tuned with the DINO framework for face anti-spoofing. The ViT (DINO) model consistently outperformed the EfficientNet b2 model across all key metrics, indicating its superior ability to distinguish between genuine and spoofed faces. This means that leveraging advanced transformer-based architectures and self-supervised learning frameworks can substantially enhance the safety and dependability of biometric authentication systems. This improvement is crucial as it addresses the growing threat of spoofing attacks in various applications, from personal device security to access control in high-security environments. The findings underscore the importance of adopting cutting-edge AI technologies to safeguard biometric systems against increasingly sophisticated spoofing techniques. In summary, integrating ViT models with self-supervised learning offers a powerful solution for improving the resilience and reliability of face anti-spoofing systems.

\section{Acknowledgments}
This research was funded in part by the Science Committee of the Ministry of Science and Higher Education of the Republic of Kazakhstan (Grant No. AP22786412) and in part by the Kazakh-British Technical University.
\bibliography{reference}

\begin{thebibliography}{10}
\providecommand{\url}[1]{#1}
\csname url@samestyle\endcsname
\providecommand{\newblock}{\relax}
\providecommand{\bibinfo}[2]{#2}
\providecommand{\BIBentrySTDinterwordspacing}{\spaceskip=0pt\relax}
\providecommand{\BIBentryALTinterwordstretchfactor}{4}
\providecommand{\BIBentryALTinterwordspacing}{\spaceskip=\fontdimen2\font plus
\BIBentryALTinterwordstretchfactor\fontdimen3\font minus \fontdimen4\font\relax}
\providecommand{\BIBforeignlanguage}[2]{{%
\expandafter\ifx\csname l@#1\endcsname\relax
\typeout{** WARNING: IEEEtran.bst: No hyphenation pattern has been}%
\typeout{** loaded for the language `#1'. Using the pattern for}%
\typeout{** the default language instead.}%
\else
\language=\csname l@#1\endcsname
\fi
#2}}
\providecommand{\BIBdecl}{\relax}
\BIBdecl

\bibitem{frsbio}
\BIBentryALTinterwordspacing
E.~Vazquez-Fernandez and D.~Gonzalez-Jimenez, ``Face recognition for authentication on mobile devices,'' \emph{Image and Vision Computing}, vol.~55, p. 31–33, Nov. 2016. [Online]. Available: \url{http://dx.doi.org/10.1016/j.imavis.2016.03.018}
\BIBentrySTDinterwordspacing

\bibitem{Petrescu2019}
\BIBentryALTinterwordspacing
R.~V. Petrescu, ``Face recognition as a biometric application,'' \emph{SSRN Electronic Journal}, 2019. [Online]. Available: \url{http://dx.doi.org/10.2139/ssrn.3417325}
\BIBentrySTDinterwordspacing

\bibitem{Nagesh2023}
\BIBentryALTinterwordspacing
M.~P. Nagesh, ``Face recognition systems,'' \emph{International Journal for Research in Applied Science and Engineering Technology}, vol.~11, no.~3, p. 962–964, Mar. 2023. [Online]. Available: \url{http://dx.doi.org/10.22214/ijraset.2023.49567}
\BIBentrySTDinterwordspacing

\bibitem{fai}
\BIBentryALTinterwordspacing
T.~I. Dhamecha, S.~Ghosh, M.~Vatsa, and R.~Singh, ``Kernelized heterogeneity-aware cross-view face recognition,'' \emph{Frontiers in Artificial Intelligence}, vol.~4, Jul. 2021. [Online]. Available: \url{http://dx.doi.org/10.3389/frai.2021.670538}
\BIBentrySTDinterwordspacing

\bibitem{Chowdhry2013}
\BIBentryALTinterwordspacing
D.~A. Chowdhry, A.~Hussain, M.~Z. Ur~Rehman, F.~Ahmad, A.~Ahmad, and M.~Pervaiz, ``Smart security system for sensitive area using face recognition,'' in \emph{2013 IEEE Conference on Sustainable Utilization and Development in Engineering and Technology (CSUDET)}.\hskip 1em plus 0.5em minus 0.4em\relax IEEE, May 2013. [Online]. Available: \url{http://dx.doi.org/10.1109/CSUDET.2013.6670976}
\BIBentrySTDinterwordspacing

\bibitem{AbdElaziz2021}
\BIBentryALTinterwordspacing
A.~AbdElaziz, ``A survey of smartphone-based face recognition systems for security purposes,'' \emph{Kafrelsheikh Journal of Information Sciences}, vol.~2, no.~1, p. 1–7, Aug. 2021. [Online]. Available: \url{http://dx.doi.org/10.21608/kjis.2021.5484.1006}
\BIBentrySTDinterwordspacing

\bibitem{ErdogmusNesli2014Spoofing}
{ErdogmusNesli} and {MarcelSebastien}, ``Spoofing {Face} {Recognition} {With} 3d {Masks},'' 2014.

\bibitem{Hamdan2018detection}
B.~Hamdan and K.~Mokhtar, ``The detection of spoofing by 3d mask in a 2d identity recognition system,'' \emph{Egyptian Informatics Journal}, vol.~19, no.~2, pp. 75--82, 7 2018.

\bibitem{Omar2015Evaluating}
L.~Omar and I.~Ivrissimtzis, ``Evaluating the {Resilience} of {Face} {Recognition} {Systems} {Against} {Malicious} {Attacks},'' in \emph{Procedings of the {Proceedings} of the 7th {UK} {British} {Machine} {Vision} {Workshop} 2015}.\hskip 1em plus 0.5em minus 0.4em\relax British Machine Vision Association, 2015.

\bibitem{Omar2016Designing}
------, ``Designing a {Facial} {Spoofing} {Database} for {Processed} {Image} {Attacks},'' in \emph{7th {International} {Conference} on {Imaging} for {Crime} {Detection} and {Prevention} ({ICDP} 2016)}.\hskip 1em plus 0.5em minus 0.4em\relax {Institution of Engineering and Technology}, 2016.

\bibitem{Latifah2023Anomaly}
{Latifah Abduh}, {Luma Omar}, and {I. Ivrissimtzis}, ``Anomaly {Detection} with {Transformer} in {Face} {Anti}-spoofing,'' \emph{Journal of WSCG}, 2023.

\bibitem{Chen2023Domain}
{Chen-Hao Liao}, {Wen-Cheng Chen}, {Hsuan-Tung Liu}, {Yi-Ren Yeh}, {Min-Chun Hu}, and {Chu-Song Chen}, ``Domain {Invariant} {Vision} {Transformer} {Learning} for {Face} {Anti}-spoofing,'' \emph{IEEE Workshop/Winter Conference on Applications of Computer Vision}, 2023.

\bibitem{Liu2023}
\BIBentryALTinterwordspacing
A.~Liu, Z.~Tan, Z.~Yu, C.~Zhao, J.~Wan, Y.~Liang, Z.~Lei, D.~Zhang, S.~Z. Li, and G.~Guo, ``Fm-vit: Flexible modal vision transformers for face anti-spoofing,'' \emph{IEEE Transactions on Information Forensics and Security}, vol.~18, p. 4775–4786, 2023. [Online]. Available: \url{http://dx.doi.org/10.1109/TIFS.2023.3296330}
\BIBentrySTDinterwordspacing

\bibitem{Liao2023Domain}
C.-H. Liao, W.-C. Chen, H.-T. Liu, Y.-R. Yeh, M.-C. Hu, and C.-S. Chen, ``Domain {Invariant} {Vision} {Transformer} {Learning} for {Face} {Anti}-spoofing,'' in \emph{2023 {IEEE}/{CVF} {Winter} {Conference} on {Applications} of {Computer} {Vision} ({WACV})}.\hskip 1em plus 0.5em minus 0.4em\relax IEEE, 1 2023.

\bibitem{Lee2023Robust}
Y.~Lee, Y.~Kwak, and J.~Shin, ``Robust {Face} {Anti}-{Spoofing} {Framework} with {Convolutional} {Vision} {Transformer},'' in \emph{2023 {IEEE} {International} {Conference} on {Image} {Processing} ({ICIP})}.\hskip 1em plus 0.5em minus 0.4em\relax IEEE, oct 8 2023.

\bibitem{7748511}
Z.~Boulkenafet, J.~Komulainen, and A.~Hadid, ``Face antispoofing using speeded-up robust features and fisher vector encoding,'' \emph{IEEE Signal Processing Letters}, vol.~24, no.~2, pp. 141--145, 2017.

\bibitem{7487030}
K.~Patel, H.~Han, and A.~K. Jain, ``Secure face unlock: Spoof detection on smartphones,'' \emph{IEEE Transactions on Information Forensics and Security}, vol.~11, no.~10, pp. 2268--2283, 2016.

\bibitem{Agarwal2016Face}
A.~Agarwal, R.~Singh, and M.~Vatsa, ``Face anti-spoofing using {Haralick} features,'' in \emph{2016 {IEEE} 8th {International} {Conference} on {Biometrics} {Theory}, {Applications} and {Systems} ({BTAS})}.\hskip 1em plus 0.5em minus 0.4em\relax IEEE, 9 2016.

\bibitem{Fourati2017Face}
E.~Fourati, W.~Elloumi, and A.~Chetouani, ``Face anti-spoofing with image quality assessment,'' in \emph{2017 2nd {International} {Conference} on {Bio}-engineering for {Smart} {Technologies} ({BioSMART})}.\hskip 1em plus 0.5em minus 0.4em\relax IEEE, 8 2017.

\bibitem{Atoum2017Face}
Y.~Atoum, Y.~Liu, A.~Jourabloo, and X.~Liu, ``Face anti-spoofing using patch and depth-based {CNNs},'' in \emph{2017 {IEEE} {International} {Joint} {Conference} on {Biometrics} ({IJCB})}.\hskip 1em plus 0.5em minus 0.4em\relax IEEE, 10 2017.

\bibitem{Siddiqui2016}
T.~A. Siddiqui, S.~Bharadwaj, T.~I. Dhamecha, A.~Agarwal, M.~Vatsa, R.~Singh, and N.~Ratha, ``Face anti-spoofing with multifeature videolet aggregation,'' 2016, pp. 1035--1040.

\bibitem{Galbally2014Face}
J.~Galbally and S.~Marcel, ``Face {Anti}-spoofing {Based} on {General} {Image} {Quality} {Assessment},'' in \emph{2014 22nd {International} {Conference} on {Pattern} {Recognition}}.\hskip 1em plus 0.5em minus 0.4em\relax IEEE, 8 2014.

\bibitem{Boulkenafet2015Face}
Z.~Boulkenafet, J.~Komulainen, and A.~Hadid, ``Face anti-spoofing based on color texture analysis,'' in \emph{2015 {IEEE} {International} {Conference} on {Image} {Processing} ({ICIP})}.\hskip 1em plus 0.5em minus 0.4em\relax IEEE, 9 2015.

\bibitem{XiaobaiLi2016}
\BIBentryALTinterwordspacing
X.~Li, J.~Komulainen, G.~Zhao, P.-C. Yuen, and M.~Pietikainen, ``Generalized face anti-spoofing by detecting pulse from face videos,'' in \emph{2016 23rd International Conference on Pattern Recognition (ICPR)}.\hskip 1em plus 0.5em minus 0.4em\relax IEEE, Dec. 2016. [Online]. Available: \url{http://dx.doi.org/10.1109/ICPR.2016.7900300}
\BIBentrySTDinterwordspacing

\bibitem{Aff2013RFIDAF}
\BIBentryALTinterwordspacing
A.~Aff, M.~Awedh, and M.~H.~A. Alghamdi, ``Rfid and face recognition based security and access control system,'' \emph{International Journal of Innovative Research in Science, Engineering and Technology}, vol.~2, pp. 5955--5964, 2013. [Online]. Available: \url{https://api.semanticscholar.org/CorpusID:13542387}
\BIBentrySTDinterwordspacing

\bibitem{9070853}
S.~Garg, S.~Mittal, P.~Kumar, and V.~Anant~Athavale, ``Debnet: Multilayer deep network for liveness detection in face recognition system,'' in \emph{2020 7th International Conference on Signal Processing and Integrated Networks (SPIN)}, 2020, pp. 1136--1141.

\bibitem{9112543}
S.~Jafri, S.~Chawan, and A.~Khan, ``Face recognition using deep neural network with "livenessnet",'' in \emph{2020 International Conference on Inventive Computation Technologies (ICICT)}, 2020, pp. 145--148.

\bibitem{Wang2022Face}
Z.~Wang, Q.~Wang, W.~Deng, and G.~Guo, ``Face {Anti}-{Spoofing} {Using} {Transformers} {With} {Relation}-{Aware} {Mechanism},'' \emph{IEEE Transactions on Biometrics, Behavior, and Identity Science}, vol.~4, no.~3, pp. 439--450, 7 2022.

\bibitem{Tianqi2021Transformer}
{Tianqi Zhou}, ``Transformer-based {Face} {Anti}-{Spoofing} {Algorithm},'' 2021.

\bibitem{Hsin2022Adaptive}
{Hsin-Ping Huang}, {Deqing Sun}, {Yaojie Liu}, {Wen-Sheng Chu}, {Taihong Xiao}, {Jinwei Yuan}, {Hartwig Adam}, and {Ming-Hsuan Yang}, ``Adaptive {Transformers} for {Robust} {Few}-shot {Cross}-domain {Face} {Anti}-spoofing,'' \emph{European Conference on Computer Vision}, 2022.

\bibitem{George2021}
A.~George and S.~Marcel, ``On the effectiveness of vision transformers for zero-shot face anti-spoofing,'' 2021, pp. 1--8.

\bibitem{unsuperv}
Y.~Liu, Y.~Chen, M.~Gou, C.-T. Huang, Y.~Wang, W.~Dai, and H.~Xiong, ``Towards unsupervised domain generalization for face anti-spoofing,'' in \emph{2023 IEEE/CVF International Conference on Computer Vision (ICCV)}, 2023, pp. 20\,597--20\,607.

\bibitem{huang}
\BIBentryALTinterwordspacing
H.-P. Huang, D.~Sun, Y.~Liu, W.-S. Chu, T.~Xiao, J.~Yuan, H.~Adam, and M.-H. Yang, ``Adaptive transformers for robust few-shot cross-domain face anti-spoofing,'' 2022. [Online]. Available: \url{https://arxiv.org/abs/2203.12175}
\BIBentrySTDinterwordspacing

\bibitem{9484333}
A.~George and S.~Marcel, ``On the effectiveness of vision transformers for zero-shot face anti-spoofing,'' in \emph{2021 IEEE International Joint Conference on Biometrics (IJCB)}, 2021, pp. 1--8.

\bibitem{9817442}
Z.~Wang, Q.~Wang, W.~Deng, and G.~Guo, ``Face anti-spoofing using transformers with relation-aware mechanism,'' \emph{IEEE Transactions on Biometrics, Behavior, and Identity Science}, vol.~4, no.~3, pp. 439--450, 2022.

\bibitem{zhang2020celebaspoof}
Y.~Zhang, Z.~Yin, Y.~Li, G.~Yin, J.~Yan, J.~Shao, and Z.~Liu, ``Celeba-spoof: Large-scale face anti-spoofing dataset with rich annotations,'' 2020.

\bibitem{zhang2020casiasurf}
S.~Zhang, A.~Liu, J.~Wan, Y.~Liang, G.~Guo, S.~Escalera, H.~J. Escalante, and S.~Z. Li, ``Casia-surf: A large-scale multi-modal benchmark for face anti-spoofing,'' 2020.

\bibitem{fai2}
\BIBentryALTinterwordspacing
N.~Ilinykh and S.~Dobnik, ``What does a language-and-vision transformer see: The impact of semantic information on visual representations,'' \emph{Frontiers in Artificial Intelligence}, vol.~4, Dec. 2021. [Online]. Available: \url{http://dx.doi.org/10.3389/frai.2021.767971}
\BIBentrySTDinterwordspacing

\bibitem{Dosovitskiy2020}
\BIBentryALTinterwordspacing
A.~Dosovitskiy, L.~Beyer, A.~Kolesnikov, D.~Weissenborn, X.~Zhai, T.~Unterthiner, M.~Dehghani, M.~Minderer, G.~Heigold, S.~Gelly, J.~Uszkoreit, and N.~Houlsby, ``An image is worth 16x16 words: Transformers for image recognition at scale,'' \emph{CoRR}, vol. abs/2010.11929, 2020. [Online]. Available: \url{https://arxiv.org/abs/2010.11929}
\BIBentrySTDinterwordspacing

\bibitem{Caron2021}
\BIBentryALTinterwordspacing
M.~Caron, H.~Touvron, I.~Misra, H.~Jégou, J.~Mairal, P.~Bojanowski, and A.~Joulin, ``Emerging properties in self-supervised vision transformers,'' \emph{CoRR}, vol. abs/2104.14294, 2021. [Online]. Available: \url{https://arxiv.org/abs/2104.14294}
\BIBentrySTDinterwordspacing

\bibitem{tan2020efficientnet}
M.~Tan and Q.~V. Le, ``Efficientnet: Rethinking model scaling for convolutional neural networks,'' 2020.

\bibitem{xie2020selftraining}
Q.~Xie, M.-T. Luong, E.~Hovy, and Q.~V. Le, ``Self-training with noisy student improves imagenet classification,'' 2020.

\bibitem{ISO30107-3}
{International Organization for Standardization}, ``{Information technology — Biometric presentation attack detection — Part 3: Testing and reporting},'' ISO/IEC 30107-3:2023, 2023, available: https://www.iso.org/standard/79520.html.

\bibitem{dis1}
M.~Marais, D.~Brown, J.~Connan, and A.~Boby, ``Facial liveness and anti-spoofing detection using vision transformers,'' in \emph{Southern Africa Telecommunication Networks and Applications Conference (SATNAC) 2023}, 08 2023.

\bibitem{Huang2022Adaptive}
H.-P. Huang, D.~Sun, Y.~Liu, W.-S. Chu, T.~Xiao, J.~Yuan, H.~Adam, and M.-H. Yang, ``Adaptive {Transformers} for {Robust} {Few}-shot {Cross}-domain {Face} {Anti}-spoofing,'' 2022.

\bibitem{Yang2024}
\BIBentryALTinterwordspacing
J.~Yang, F.~Chen, R.~K. Das, Z.~Zhu, and S.~Zhang, ``Adaptive-avg-pooling based attention vision transformer for face anti-spoofing,'' in \emph{ICASSP 2024 - 2024 IEEE International Conference on Acoustics, Speech and Signal Processing (ICASSP)}.\hskip 1em plus 0.5em minus 0.4em\relax IEEE, Apr. 2024. [Online]. Available: \url{http://dx.doi.org/10.1109/icassp48485.2024.10446940}
\BIBentrySTDinterwordspacing

\bibitem{George2021On}
A.~George and S.~Marcel, ``On the {Effectiveness} of {Vision} {Transformers} for {Zero}-shot {Face} {Anti}-{Spoofing},'' in \emph{2021 {IEEE} {International} {Joint} {Conference} on {Biometrics} ({IJCB})}.\hskip 1em plus 0.5em minus 0.4em\relax IEEE, aug 4 2021.

\bibitem{Orfao2023}
\BIBentryALTinterwordspacing
J.~Orfao and D.~van~der Haar, ``Keyframe and gan-based data augmentation for face anti-spoofing,'' in \emph{Proceedings of the 12th International Conference on Pattern Recognition Applications and Methods}.\hskip 1em plus 0.5em minus 0.4em\relax SCITEPRESS - Science and Technology Publications, 2023. [Online]. Available: \url{http://dx.doi.org/10.5220/0011648400003411}
\BIBentrySTDinterwordspacing

\bibitem{Watanabe2022}
\BIBentryALTinterwordspacing
K.~Watanabe, K.~Ito, and T.~Aoki, ``Spoofing attack detection in face recognition system using vision transformer with patch-wise data augmentation,'' in \emph{2022 Asia-Pacific Signal and Information Processing Association Annual Summit and Conference (APSIPA ASC)}.\hskip 1em plus 0.5em minus 0.4em\relax IEEE, Nov. 2022. [Online]. Available: \url{http://dx.doi.org/10.23919/APSIPAASC55919.2022.9979996}
\BIBentrySTDinterwordspacing

\bibitem{fan2024vitar}
Q.~Fan, Q.~You, X.~Han, Y.~Liu, Y.~Tao, H.~Huang, R.~He, and H.~Yang, ``Vitar: Vision transformer with any resolution,'' 2024.

\bibitem{fuzz-ie}
J.~C. Córdova, C.~Flores, and J.~Andreu-Perez, ``Emgtfnet: Fuzzy vision transformer to decode upperlimb semg signals for hand gestures recognition,'' in \emph{2023 IEEE International Conference on Fuzzy Systems (FUZZ)}, 2023, pp. 1--6.

\bibitem{Kozlov2024}
\BIBentryALTinterwordspacing
P.~Kozlov, A.~Akram, and P.~Shamoi, ``Fuzzy approach for audio-video emotion recognition in computer games for children,'' \emph{Procedia Computer Science}, vol. 231, p. 771–778, 2024. [Online]. Available: \url{http://dx.doi.org/10.1016/j.procs.2023.12.139}
\BIBentrySTDinterwordspacing

\end{thebibliography}
\end{document}